\documentclass[letterpaper]{article}
\usepackage[preprint]{aaai2027}
\usepackage[hyphens]{url}

\usepackage{amsmath}
\usepackage{amssymb}
\usepackage{graphicx}
\usepackage{booktabs}
\usepackage{natbib}
\usepackage{caption}

\title{Latent Thought Credit: Multi-Answer Credit Assignment for Latent Reasoning}

\author{
Xuyang Zhao\textsuperscript{\rm 1},
Liting Zhang\textsuperscript{\rm 1},
Zichen Xu\textsuperscript{\rm 1},
Yong Chen\textsuperscript{\rm 2},
Wenjia Zeng\textsuperscript{\rm 2},
Shiwan Zhao\textsuperscript{\rm 1}\corresponding,\\
Qicheng Li\textsuperscript{\rm 1}\corresponding
}
\affiliations{
\textsuperscript{\rm 1}TMCC, College of Computer Science, Nankai University, Tianjin, China\\
\textsuperscript{\rm 2}Lingxi (Beijing) Technology Co., Ltd.\\
xychao@mail.nankai.edu.cn, zhaosw@gmail.com, liqicheng@nankai.edu.cn
}

\begin{document}
\maketitle

\begin{abstract}
Latent reasoning allows language models to carry out intermediate reasoning in continuous latent representations rather than fully externalizing it as discrete chains of thought. However, assigning credit to such latent thoughts from answer-only rewards is difficult: a single final answer mixes thought quality with answer-sampling noise. We propose \textbf{Latent Thought Credit (LTC)}, a hierarchical credit-assignment framework for latent reasoning. For each prompt, LTC samples multiple latent thoughts, fixes the context after each thought, and estimates thought-level expected reward by averaging rewards over multiple answers generated from that fixed context. LTC uses thought-level advantages to optimize the latent-thought phase, answer-level advantages to optimize the answer phase, and an advantage-weighted thought-matching objective that helps the policy reproduce high-credit latent thoughts. We instantiate LTC in a GRPO-style on-policy training framework and evaluate it across mathematical reasoning and STEM multiple-choice tasks. LTC achieves the best average accuracy among the compared methods, while ablations and fixed-context diagnostics show that multi-answer estimation reduces reward-estimation error and mitigates ambiguous or incorrect thought-level credit.
\end{abstract}

\section{Introduction}

Reinforcement learning with verifiable rewards has become an effective way to improve language-model reasoning, particularly when intermediate reasoning is represented as textual chains of thought \citep{cobbe2021trainingverifierssolvemath,lightman2023letsverifystepstep,wei2023chainofthoughtpromptingelicitsreasoning,shao2024deepseekmath}. Beyond explicit text, recent latent-reasoning methods perform intermediate computation through continuous hidden states, soft tokens, or hybrid token-latent representations \citep{zhu2025surveylatentreasoning,hao2024training,su2025token,yue2025hybrid}. These approaches expand the space in which models can represent and explore intermediate reasoning.

\begin{figure}[!t]
  \centering
  \makebox[\linewidth][c]{%
    \includegraphics[width=1.03\linewidth]{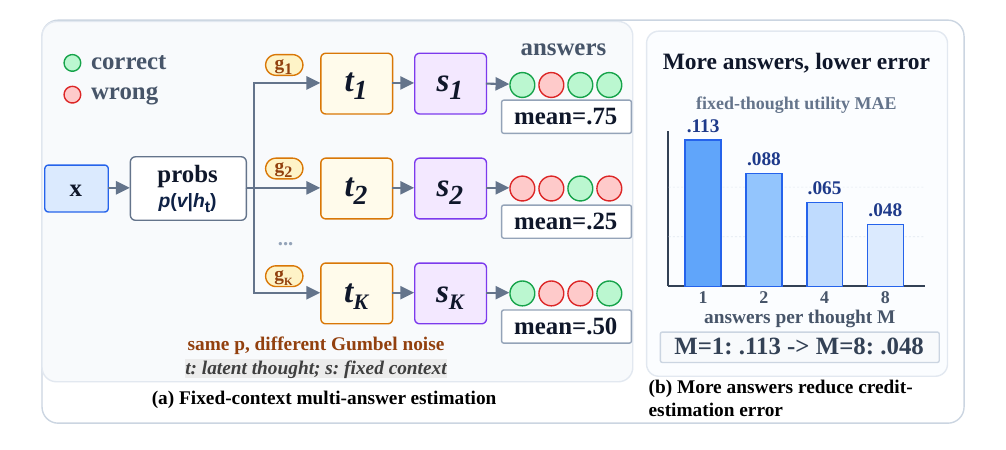}%
  }
  \caption{Fixed-context diagnostic for latent-thought credit. A single answer can conflate thought quality with answer-sampling noise; averaging more answers under the same context reduces estimation error.}
  \label{fig:fixed-context-probe-intro}
\end{figure}

However, moving reasoning from explicit token sequences into continuous latent states introduces a distinct credit-assignment challenge. A sampled latent thought is evaluated only through its downstream answers, whose rewards depend on both the quality of the thought and stochasticity in answer generation. Consequently, the reward of a single sampled answer can be a noisy estimate of the thought's utility. As illustrated in Figure \ref{fig:fixed-context-probe-intro}, fixing the post-thought context and averaging rewards across multiple answers provides a more stable estimate of the thought's expected utility.

Existing latent and soft-reasoning methods provide useful mechanisms for continuous computation, exploration, and test-time search, but they generally do not explicitly estimate the expected downstream reward of each sampled latent thought. We propose \textbf{Latent Thought Credit (LTC)}, a hierarchical credit-assignment framework for latent reasoning. For each prompt, LTC samples multiple latent thoughts, freezes the post-thought context of each thought, and samples multiple answers from that context. The mean answer reward estimates thought-level utility; LTC then uses thought-level advantages to optimize the latent-thought phase, answer-level advantages to optimize the answer phase, and an advantage-weighted thought-matching objective to help the current policy reproduce high-credit latent thoughts.

Our main contributions are as follows.
\begin{itemize}
  \item We reformulate training-signal construction for latent reasoning as thought-level reward estimation: by fixing the latent thought and averaging rewards over multiple answers, we estimate its expected reward.

  \item We propose LTC, which optimizes latent-thought generation with thought-level advantages, optimizes answer generation with answer-level advantages, and adds an advantage-weighted thought-matching objective that encourages the current policy to reproduce high-credit rollout latent thoughts.

  \item We empirically demonstrate the effectiveness of LTC on reasoning and STEM benchmarks, and further analyze its behavior through component ablations and fixed-context diagnostics, showing that multi-answer estimation stabilizes thought-level credit assignment.

\end{itemize}

\section{Related Work}

\subsection{Discrete Textual Reasoning}

Textual reasoning in LLMs is commonly improved with chain-of-thought prompting, self-consistency, and search over discrete reasoning branches \citep{wei2023chainofthoughtpromptingelicitsreasoning,wang2023selfconsistencyimproveschainthought,yao2023treethoughtsdeliberateproblem}. Verifiable rewards and GRPO-style training further optimize whole token sequences as reasoning actions \citep{cobbe2021trainingverifierssolvemath,lightman2023letsverifystepstep,shao2024deepseekmath}. Recent branch-based variants improve planning or token-wise branch merging \citep{dou2025plan,tang2026multiplex}. GRPO-MA samples multiple answer continuations from each discrete thought branch and uses their rewards to construct separate thought- and answer-level advantages \citep{wang2026treestylebranchingmattersthought}. These methods show that branch structure matters, but their credit signals remain attached to discrete or token-anchored reasoning objects rather than continuous latent thoughts.

\subsection{Continuous Latent Reasoning}

Latent reasoning methods replace or augment textual intermediate reasoning with continuous internal computation \citep{zhu2025surveylatentreasoning}. Coconut, Token Assorted, HRPO, soft-thinking methods, and latent test-time optimization explore continuous or hybrid computation for representation, sampling, search, or RL training \citep{hao2024training,su2025token,yue2025hybrid,zheng2025soft,geiping2025scalingtesttimecomputelatent,li2025seekdarkreasoningtesttime}. However, they do not directly estimate the fixed-context expected reward of a sampled latent thought. Recent latent-trajectory credit work such as RLTT targets which latent steps should receive reward \citep{williams2026prioritize}; LTC instead compares complete latent thoughts by averaging multiple downstream answers under each fixed context.

\section{Method}


\begin{figure*}[t]
  \centering
  \includegraphics[width=\textwidth]{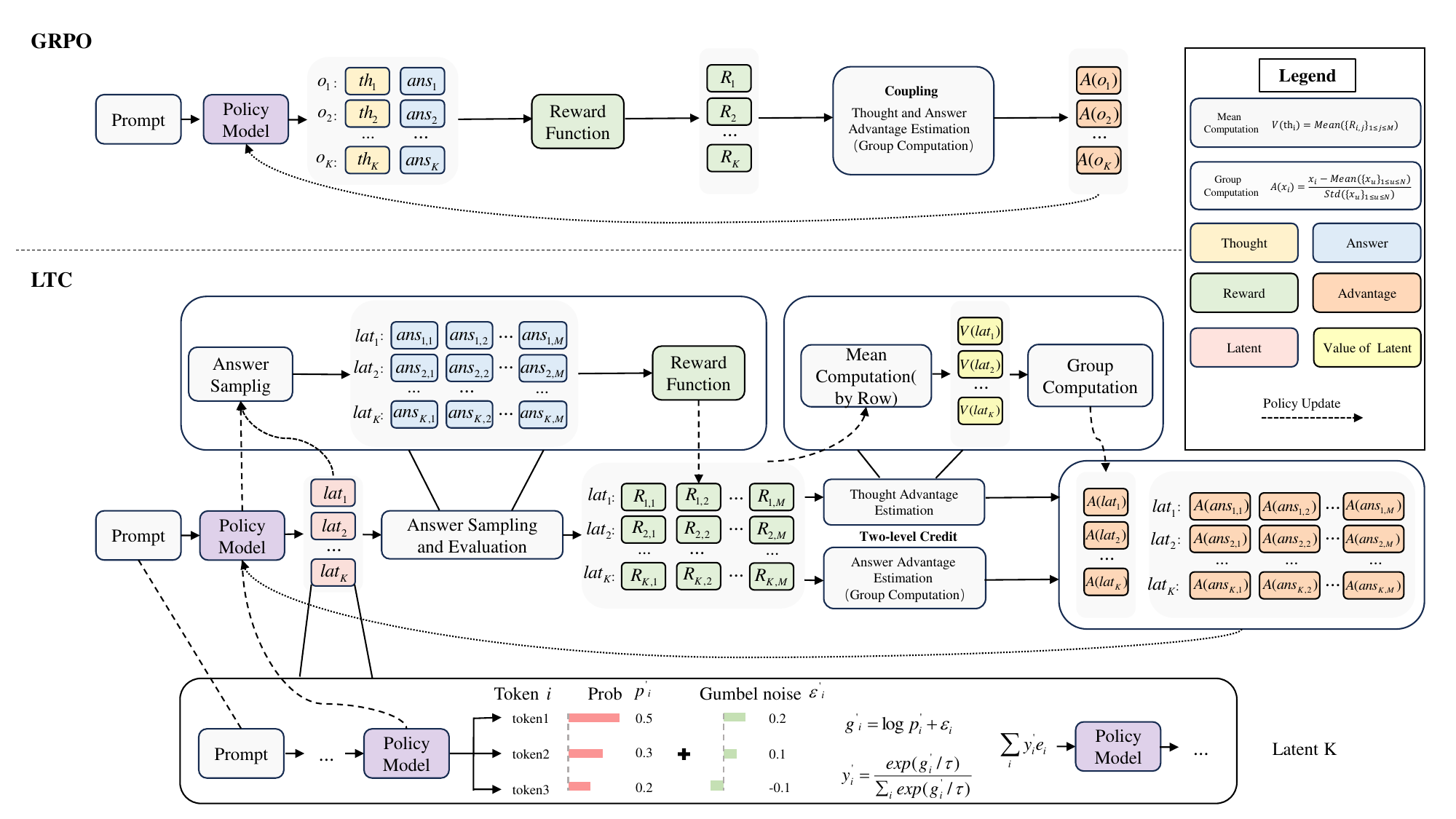}
  \caption{Overview of LTC compared with standard GRPO. GRPO uses a single group-relative advantage for the complete rollout, whereas LTC samples multiple answers for each latent thought, estimates thought-level utility from fixed-context rewards, and weights policy updates at latent-thought and answer positions with thought-level and answer-level advantages, respectively.}
  \label{fig:ltc-method}
\end{figure*}

Figure~\ref{fig:ltc-method} summarizes the LTC pipeline. The remainder of this section formalizes the latent-thought rollout, the multi-answer reward estimator, and the hierarchical training objective.

\subsection{Latent Thought Rollout}

For each prompt $x$, LTC samples $K$ latent thoughts, retains each post-thought context $s_i$, and samples $M$ discrete answers from that fixed context.

To sample the $i$-th latent thought, we inject stochasticity into the latent thought. At latent thought step $t$, let $h_{i,t}$ be the current state, let $\theta_{\mathrm{old}}$ denote the rollout policy parameters, and let $\tau_{\mathrm{think}}$ be the latent-thought sampling temperature. The rollout policy first outputs clean logits $\ell_{\theta_{\mathrm{old}}}(h_{i,t})$ before Gumbel perturbation. After adding noise $g_{i,t}$, it produces a soft distribution over the vocabulary simplex using the Gumbel-Softmax/Concrete relaxation \citep{jang2017categoricalreparameterizationgumbelsoftmax,maddison2017concretedistributioncontinuousrelaxation}:

\begin{equation}
\label{eq:gumbel-softmax-thought}
q_{i,t}=\operatorname{softmax}\left(\frac{\ell_{\theta_{\mathrm{old}}}(h_{i,t})+g_{i,t}}{\tau_{\mathrm{think}}}\right),
\end{equation}

Let $\mathcal{V}$ be the vocabulary, and let $E_v$ denote the embedding of token $v$. The latent thought token fed into the model is the embedding mixture

\begin{equation}
\label{eq:latent-thought-token}
z_{i,t}=\frac{\sum_{v\in\mathcal{V}}q_{i,t}(v)E_v}{\lVert q_{i,t}\rVert}.
\end{equation}

Here $q_{i,t}$ is the stochastic rollout distribution, and $z_{i,t}$ is the corresponding normalized latent thought token, a continuous embedding mixture rather than a discrete token. The resulting latent thought, with length $T_i$, is

\[
\tau_i^{\mathrm{think}}
=(z_{i,1},z_{i,2},\ldots,z_{i,T_i}),\quad i=1,\ldots,K.
\]

After the $i$-th thought ends, let $s_i$ denote the retained post-thought context. Keeping $s_i$ fixed, the rollout policy samples $M$ answer sequences of lengths $L_{ij}^{\mathrm{ans}}$:

\[
\begin{aligned}
y_{ij}
&=(y_{ij,1},\ldots,y_{ij,L_{ij}^{\mathrm{ans}}}),\\
y_{ij}
&\sim \pi_{\theta_{\mathrm{old}}}^{\mathrm{ans}}(\cdot\mid x,s_i),\quad j=1,\ldots,M.
\end{aligned}
\]

Each answer receives a scalar reward $r_{ij}=R(x,y_{ij})$ from a reward function.

\subsection{Thought-Level Reward Estimation}

The estimation target is the thought-level expected reward when the context $s_i$ is held fixed. For the $i$-th thought, this target can be written as

\begin{equation}
\label{eq:thought-reward-target}
\mu_i=\mathbb{E}_{y\sim\pi_{\theta_{\mathrm{old}}}^{\mathrm{ans}}(\cdot\mid x,s_i)}\left[R(x,y)\right].
\end{equation}

Since $\mu_i$ is not directly observable, we sample $M$ answers under the same context $s_i$ and estimate it by the average answer reward:

\begin{equation}
\label{eq:thought-reward-estimator}
\hat{\mu}_i=\frac{1}{M}\sum_{j=1}^{M}r_{ij}.
\end{equation}

\subsection{Hierarchical Credit Construction}

Given $\hat{\mu}_i$, LTC constructs advantages at the thought and answer levels. The thought-level advantage compares the estimated rewards of different latent thoughts, while the answer-level advantage follows group-relative normalization over all answers sampled for the same prompt. This retains multi-answer estimation for latent-thought credit while providing a group-relative training signal for answer generation.

The thought-level advantage and answer-level advantage are defined as

\begin{equation}
\label{eq:hierarchical-advantages}
\begin{aligned}
A_i^{\mathrm{think}}
&=\frac{\hat{\mu}_i-\operatorname{mean}_{i'}(\hat{\mu}_{i'})}{\operatorname{std}_{i'}(\hat{\mu}_{i'})+\epsilon},\\
A_{ij}^{\mathrm{ans}}
&=\frac{r_{ij}-\operatorname{mean}_{i',j'}(r_{i'j'})}{\operatorname{std}_{i',j'}(r_{i'j'})+\epsilon}.
\end{aligned}
\end{equation}

The $i'$ statistics in the thought-level advantage are computed over the $K$ thoughts for the same prompt, while the joint $i',j'$ statistics in the answer-level advantage are computed over all $KM$ answers for that prompt. The constant $\epsilon$ ensures numerical stability.

\subsection{Hierarchical Policy Objective}
\label{sec:hierarchical-policy-objective}

The policy objective is computed on complete rollouts. The key change from standard GRPO is that latent-thought positions use the multi-answer thought-level advantage, whereas answer positions use the group-relative answer advantage. Let $c_{ij,1:L_{ij}}$ be the $(i,j)$-th complete rollout sequence, including latent thought tokens and discrete tokens. For notational compactness, let its policy log-probability term under the current policy parameters $\theta$ be

\begin{equation}
\label{eq:policy-log-prob}
\ell_{ij,t}(\theta)=\log\pi_\theta(c_{ij,t}\mid x,c_{ij,<t}).
\end{equation}

For latent thought positions, this term denotes the Gumbel-Softmax latent-thought policy surrogate; for answer positions, it reduces to the standard answer-token log probability.

Let $\mathcal{P}_{i}^{\mathrm{think}}$ denote the latent thought positions in the $i$-th sampled thought, and let $\mathcal{P}_{ij}^{\mathrm{ans}}$ denote the answer positions in the $(i,j)$-th rollout. Write $T_i=|\mathcal{P}_{i}^{\mathrm{think}}|$ and $U_{ij}=|\mathcal{P}_{ij}^{\mathrm{ans}}|$. Latent-thought positions are weighted by the thought-level advantage, while answer positions are weighted by the answer-level advantage. Let $\operatorname{sg}(\cdot)$ denote stop-gradient. The policy loss is decomposed as

\begin{equation}
\label{eq:policy-loss-decomposition}
\mathcal{L}_{\mathrm{policy}}
=
\mathcal{L}_{\mathrm{think}}
+
\mathcal{L}_{\mathrm{ans}}.
\end{equation}

\begin{equation}
\label{eq:hierarchical-policy-losses}
\begin{aligned}
\mathcal{L}_{\mathrm{think}}
&=
-\frac{1}{K}
\sum_{i=1}^{K}
\frac{1}{T_i} \\
&\quad
\sum_{t\in\mathcal{P}_{i}^{\mathrm{think}}}
\operatorname{sg}\left(A_i^{\mathrm{think}}\right)
\ell_{i,t}^{\mathrm{think}}(\theta),\\
\mathcal{L}_{\mathrm{ans}}
&=
-\frac{1}{KM}
\sum_{i=1}^{K}\sum_{j=1}^{M}
\frac{1}{U_{ij}} \\
&\quad
\sum_{t\in\mathcal{P}_{ij}^{\mathrm{ans}}}
\operatorname{sg}\left(A_{ij}^{\mathrm{ans}}\right)
\ell_{ij,t}^{\mathrm{ans}}(\theta).
\end{aligned}
\end{equation}

Here $\ell_{i,t}^{\mathrm{think}}$ and $\ell_{ij,t}^{\mathrm{ans}}$ are the corresponding thought-side and answer-side instances of $\ell_{ij,t}$. This loss performs advantage-weighted updates on sampled latent thought tokens and answer tokens. This objective follows a simple REINFORCE-style formulation. The main distinction of LTC is therefore not a change in the underlying update form, but a change in how credit is estimated and assigned in latent reasoning.

\subsection{Thought Matching Auxiliary}
\label{sec:thought-matching-auxiliary}

In addition to the hierarchical policy loss, we introduce a thought-matching auxiliary objective to align the current policy's latent thoughts with high-credit latent thoughts sampled by the rollout policy.

For the $i$-th sampled thought, let $z_{i,t}^{\mathrm{roll}}$ denote its rollout latent thought token at step $t$, and let $\ell_{i,t}$ denote the current policy's clean logits at the same step. Let $\mathcal{T}_i$ be the set of latent thought steps in this thought. For each $t\in\mathcal{T}_i$, we convert the clean logits into a top-$k$ embedding prediction:

\begin{equation}
\label{eq:topk-embedding-prediction}
\hat z_{i,t}
=
\sum_{v\in S_{i,t}}
\frac{\exp \ell_{i,t}(v)}
{\sum_{u\in S_{i,t}}\exp \ell_{i,t}(u)}
E_v,
\end{equation}

where $S_{i,t}$ is the top-$k$ token set under the clean logits and $E_v$ is the embedding of token $v$.

We then measure how well this clean prediction matches the rollout latent thought:

\begin{equation}
\label{eq:matching-distance}
D_i^{\mathrm{match}}
=
\frac{1}{|\mathcal{T}_i|}
\sum_{t\in\mathcal{T}_i}
\frac{
\left\|\hat z_{i,t}
-
\operatorname{sg}\left(z_{i,t}^{\mathrm{roll}}\right)
\right\|_2^2
}{d_{\mathrm{model}}}.
\end{equation}

Here $d_{\mathrm{model}}$ is the model hidden dimension used to normalize the squared embedding distance.

Different thoughts are weighted according to their thought-level advantages:

\begin{equation}
\label{eq:thought-matching-weights}
\omega_i=\frac{\exp\left( A^{\mathrm{think}}_i\right)}{\sum_{i'=1}^{K}\exp\left( A^{\mathrm{think}}_{i'}\right)}.
\end{equation}

The matching auxiliary objective is therefore

\begin{equation}
\label{eq:thought-matching-loss}
\mathcal{L}_{\mathrm{match}}=\sum_{i=1}^{K}\omega_iD_i^{\mathrm{match}}.
\end{equation}

Intuitively, Gumbel exploration produces diverse latent thoughts, multi-answer evaluation estimates the expected reward of each thought, and $\mathcal{L}_{\mathrm{match}}$ makes it easier for the policy to reproduce the soft-thinking embedding geometry of high-credit latent thoughts.

\subsection{Overall Objective and Training Procedure}

The final training objective is

\begin{equation}
\label{eq:overall-objective}
\mathcal{L}=\mathcal{L}_{\mathrm{policy}}+\lambda\mathcal{L}_{\mathrm{match}}.
\end{equation}

$\mathcal{L}_{\mathrm{policy}}$ is the hierarchical policy objective in Section~\ref{sec:hierarchical-policy-objective} and includes updates in both the soft-thinking phase and the answer phase; $\mathcal{L}_{\mathrm{match}}$ is the additional thought-matching auxiliary objective in Section~\ref{sec:thought-matching-auxiliary}; and $\lambda$ is the weight of this auxiliary objective.

\section{Experiments}

We first describe the implementation details, including training and testing settings and the main baselines. We then report the main comparison, and use ablation experiments to analyze the contributions of multi-answer thought-level reward estimation, hierarchical thought credit assignment, and the thought-matching auxiliary objective.

\subsection{Implementation Details}

\subsubsection{Training and Testing Settings}

The experiments cover three types of tasks. GSM8K evaluates basic mathematical word problems and short-chain arithmetic reasoning \citep{cobbe2021trainingverifierssolvemath}; MATH and MATH500 evaluate more challenging mathematical reasoning and held-out generalization \citep{hendrycks2021measuringmathematicalproblemsolving,lightman2023letsverifystepstep}; MMLU-STEM and ARC-Challenge (ARC-C) evaluate transfer to STEM multiple-choice understanding and knowledge reasoning \citep{hendrycks2021measuringmassivemultitasklanguage,clark2018thinksolvedquestionanswering}. Training uses the corresponding GSM8K and MATH training sets, as well as a multiple-choice training set composed of MMLU and ARC-C samples.

To ensure fair comparisons, the main experiments compare all methods under the same model, prompt format, reward computation, sampling temperature, and rollout budget. Experiments are based on the Qwen2.5-Instruct model family \citep{qwen2024qwen25technicalreport} and a GRPO-style training framework. Unless otherwise specified, LTC fixes the rollout budget to $B=K\times M=8$, and the main setting uses $(K,M)=(2,4)$. Evaluation uses greedy decoding. 

\subsubsection{Baselines}

We compare LTC with three baselines. HRPO controls for the effect of the latent-reasoning architecture itself. GRPO is a flat group-relative optimization baseline under the same budget. GRPO with Multi-Answer (GRPO-MA) samples multiple answer continuations from each discrete textual thought prefix. It computes thought-level advantages by normalizing the mean reward of each prefix and answer-level advantages by normalizing rewards across all answers sampled for the prompt \citep{wang2026treestylebranchingmattersthought}.

\subsection{Main Results}
\begin{table*}[t]
  \centering
  \small
  \renewcommand{\arraystretch}{1.08}
  \setlength{\tabcolsep}{5pt}
  \begin{tabular}{lcccccc}
    \toprule
    Method & GSM8K & MATH & MATH500 & MMLU-STEM & ARC-C & Average \\
    \midrule
    \multicolumn{7}{c}{Qwen2.5-3B-Instruct} \\
    \midrule
    GRPO & 82.71\% & 57.60\% & 57.00\% & 63.11\% & 81.83\% & 68.45\% \\
    GRPO-MA & 83.24\% & 54.60\% & 58.20\% & 64.16\% & 82.76\% & 68.59\% \\
    HRPO & 83.32\% & 57.70\% & 58.40\% & 63.97\% & 83.02\% & 69.28\% \\
    LTC & \textbf{84.46\%} & \textbf{58.40\%} & \textbf{58.80\%} & \textbf{64.86\%} & \textbf{84.30\%} & \textbf{70.16\%} \\
    \midrule
    \multicolumn{7}{c}{Qwen2.5-7B-Instruct} \\
    \midrule
    GRPO & 88.17\% & 61.80\% & 62.60\% & 68.70\% & 61.09\% & 68.47\% \\
    GRPO-MA & 87.65\% & 63.22\% & 62.20\% & 69.17\% & 75.85\% & 71.62\% \\
    HRPO & 88.55\% & \textbf{67.40\%} & 65.20\% & 65.27\% & \textbf{84.00\%} & 74.08\% \\
    LTC & \textbf{89.15\%} & 67.22\% & \textbf{67.60\%} & \textbf{70.10\%} & 82.50\% & \textbf{75.31\%} \\
    \bottomrule
  \end{tabular}
  \caption{Main experimental results across reasoning and STEM benchmarks. All values are accuracies (\%). Bold indicates the best result within each model group and column. Average is the mean over the five evaluation sets.}
  \label{tab:main-results}
\end{table*}

Table \ref{tab:main-results} reports the comparison results. In the displayed 3B results, LTC has the highest value on all five evaluation sets and the highest Average. In the displayed 7B results, LTC has the highest Average, but the per-task behavior is mixed.

By task, LTC obtains the best result within the 7B group on GSM8K, MATH500, and MMLU-STEM. On MATH, it is close to HRPO (67.22\% vs. 67.40\%), while on ARC-C it remains below HRPO (82.50\% vs. 84.00\%). 
GRPO-MA applies multi-answer evaluation to discrete token-level thought branches, whereas LTC estimates and assigns credit to continuous latent thoughts and additionally uses the thought-matching objective.

\begin{figure}[t]
  \centering
  \includegraphics[width=\linewidth]{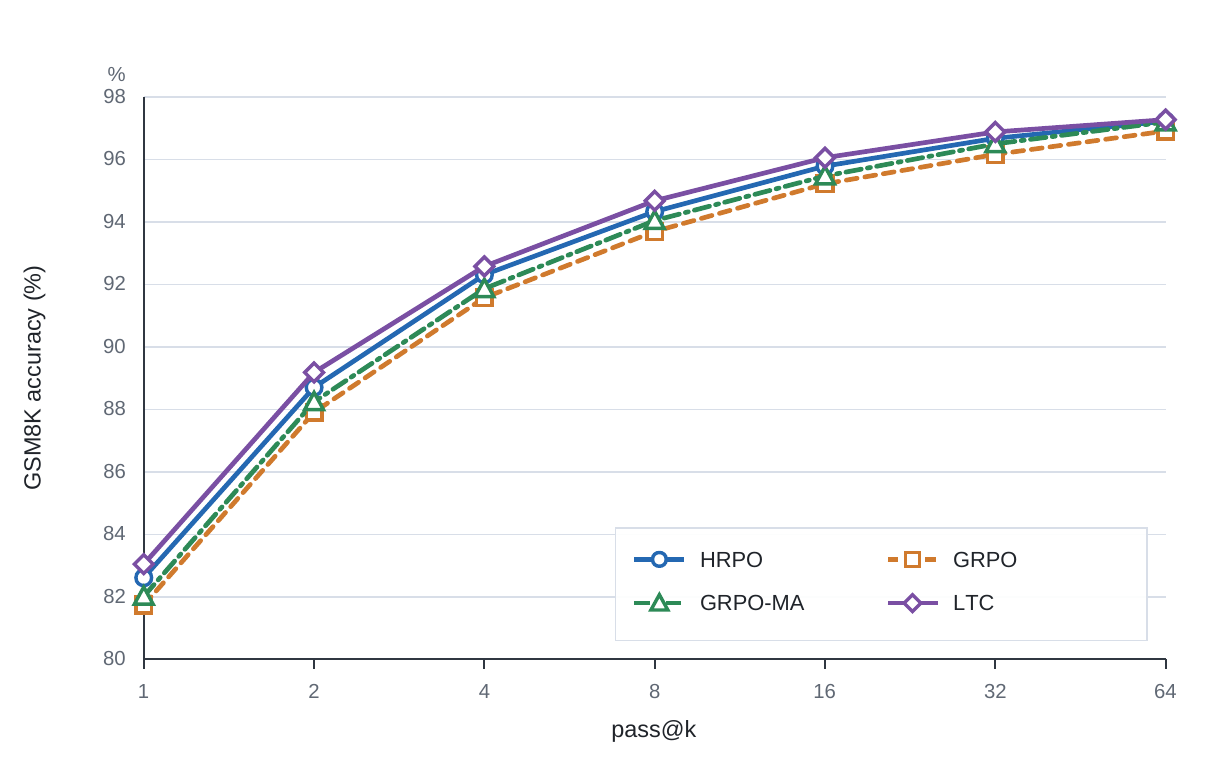}
  \caption{Qwen2.5-3B-Instruct GSM8K pass@$k$ accuracy under stochastic sampling, estimated from 64 completions per test question for each method.}
  \label{fig:gsm8k-passk-by-method}
\end{figure}

We also compare stochastic sampling performance on GSM8K for Qwen2.5-3B-Instruct using pass@$k$ estimates computed from 64 sampled completions per test question \citep{chen2021evaluatinglargelanguagemodels}. Figure \ref{fig:gsm8k-passk-by-method} shows that LTC achieves the strongest sampling accuracy across the full pass@$k$ range, with especially clear gains at low and moderate sampling budgets. The advantage narrows as $k$ increases, but LTC still remains the best-performing method at pass@64.

\subsection{Ablation Experiments}

Table \ref{tab:ablation-plan} evaluates four LTC components under identical task-specific settings. \emph{w/o latent thought} replaces continuous embedding mixtures with discrete token sequences and consequently removes thought matching; \emph{w/o Gumbel noise} sets the noise scale to zero; \emph{w/o hierarchical} replaces the nested $K\times M$ credit scheme with flat GRPO advantages over complete rollouts; and \emph{w/o thought-matching} sets $\lambda=0$.

Full LTC performs best on both datasets, reaching 84.46\% accuracy on GSM8K and 58.40\% on MATH, while every ablated variant performs worse. The largest reduction on GSM8K occurs without thought matching ($-3.11$ points), whereas removing the latent thought causes the largest reduction on MATH ($-3.80$ points). Removing Gumbel noise also consistently degrades performance, by 2.12 points on GSM8K and 1.70 points on MATH. These results support the components as a combined system, while the task-dependent ordering indicates that their relative contributions are not uniform across datasets.

\begin{table}[t]
  \centering
  \small
  \renewcommand{\arraystretch}{1.08}
  \setlength{\tabcolsep}{2.5pt}
  \begin{tabular}{p{0.40\linewidth}rrrr}
    \toprule
    & \multicolumn{2}{c}{GSM8K} & \multicolumn{2}{c}{MATH} \\
    \cmidrule(lr){2-3}\cmidrule(lr){4-5}
    Variant & Acc. & $\Delta$ & Acc. & $\Delta$ \\
    \midrule
    Full LTC & 84.46\% & -- & 58.40\% & -- \\
    w/o latent thought & 83.24\% & -1.22 & 54.60\% & -3.80 \\
    w/o Gumbel noise & 82.34\% & -2.12 & 56.70\% & -1.70 \\
    w/o hierarchical & 81.55\% & -2.91 & 56.10\% & -2.30 \\
    w/o thought-matching & 81.35\% & -3.11 & 57.20\% & -1.20 \\
    \bottomrule
  \end{tabular}
  \caption{Component ablations on GSM8K and MATH. Acc. is the test accuracy, and $\Delta$ is the change in percentage points relative to full LTC on the same task.}
  \label{tab:ablation-plan}
\end{table}

\subsection{Multi-Answer Sampling and Rollout Allocation}
\label{sec:rollout-budget-allocation}

In addition to component ablations, we examine both the number of sampled answers per latent thought and the allocation of the rollout budget between latent thoughts and answers.
Table \ref{tab:km-budget-allocation} reports GSM8K accuracy for LTC under several $(K,M)$ configurations.
Rows with $K=2$ vary the number of answers per thought, while configurations sharing $B=K\times M$ compare different allocations under the same rollout budget.

\begin{table}[t]
  \centering
  \small
  \renewcommand{\arraystretch}{1.08}
  \setlength{\tabcolsep}{5pt}
  \begin{tabular}{cccc}
    \toprule
    Budget $B$ & Thoughts $K$ & Answers $M$ & Acc. \\
    \midrule
    2 & 2 & 1 & 82.27\% \\
    4 & 2 & 2 & 82.59\% \\
    \midrule
    8
      & \begin{tabular}[c]{@{}c@{}}2\\4\end{tabular}
      & \begin{tabular}[c]{@{}c@{}}4\\2\end{tabular}
      & \begin{tabular}[c]{@{}r@{}}84.46\%\\82.49\%\end{tabular} \\
    \midrule
    16
      & \begin{tabular}[c]{@{}c@{}}2\\4\\8\end{tabular}
      & \begin{tabular}[c]{@{}c@{}}8\\4\\2\end{tabular}
      & \begin{tabular}[c]{@{}r@{}}84.52\%\\85.15\%\\83.10\%\end{tabular} \\
    \bottomrule
  \end{tabular}
  \caption{GSM8K accuracy under different allocations of latent thoughts $K$ and answers per thought $M$. Rows with $K=2$ examine the effect of increasing the number of answers per thought, while configurations sharing the same rollout budget $B=K\times M$ compare different allocations between thought breadth and answer replication. Acc. denotes the best observed checkpoint accuracy.}
  \label{tab:km-budget-allocation}
\end{table}

Among the completed $K=2$ configurations, accuracy improves from 82.27\% with one answer per thought to 84.46\% with four answers, while increasing to eight answers yields only a marginal additional gain of 0.06 percentage points.
At $B=8$, $(K,M)=(2,4)$ outperforms $(4,2)$ by 1.97 percentage points, indicating that allocating more of the budget to answer replication is more effective than sampling more latent thoughts in this setting.
At $B=16$, the balanced $(4,4)$ allocation performs best at 85.15\%, compared with 84.52\% for $(2,8)$ and 83.10\% for $(8,2)$.
Overall, these results suggest that multi-answer estimation is beneficial, but the preferred balance between thought breadth and answer replication depends on the available rollout budget; the main $(2,4)$ setting captures most of the gain at a lower budget.

\subsection{Thought Matching Hyperparameter Sensitivity}

\begin{figure}[t]
  \centering
  \includegraphics[width=\linewidth]{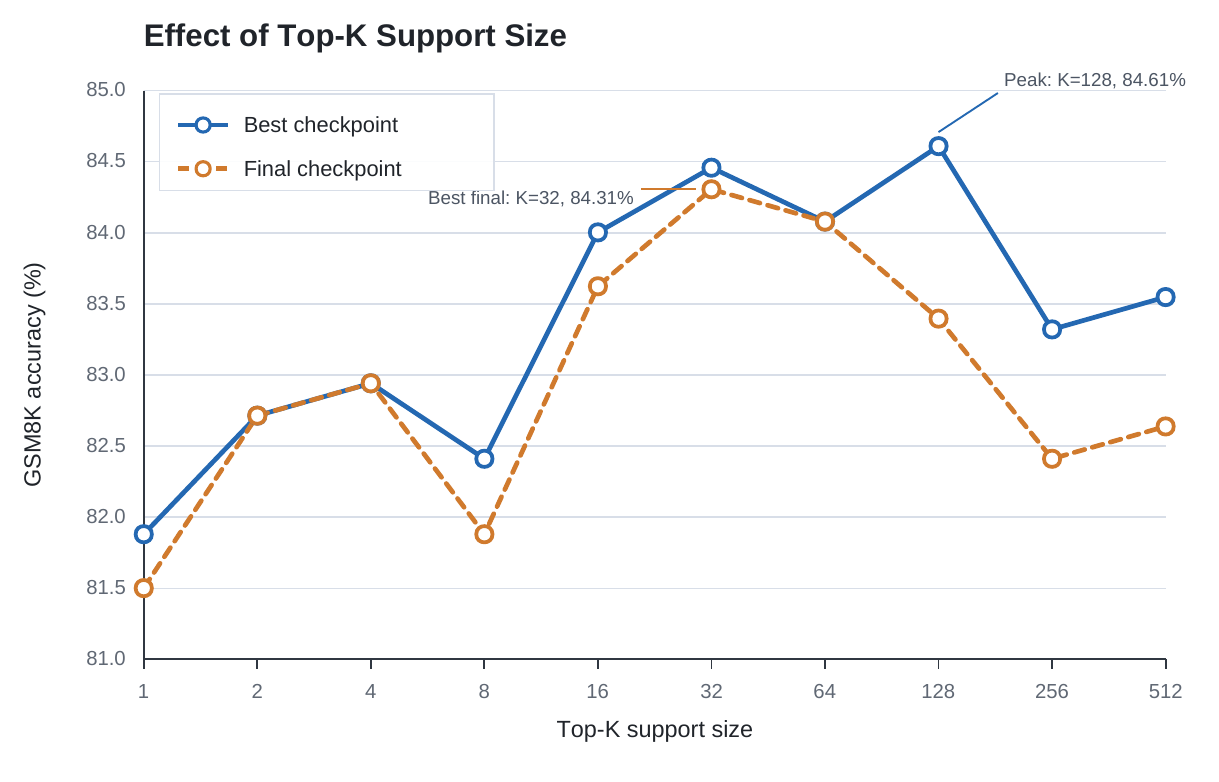}
  \caption{GSM8K sensitivity to the top-$k$ support size in the thought-matching auxiliary objective. Best and final denote peak and last-checkpoint accuracy, respectively.}
  \label{fig:topk-support-sensitivity}
\end{figure}

We examine sensitivity to the top-$k$ support size and matching strength $\lambda$. Figure \ref{fig:topk-support-sensitivity} shows a non-monotonic trend: moderate supports ($k=32$--$64$) give the strongest final results. Although $k=128$ reaches the highest peak accuracy of 84.61\%, it declines to 83.40\% at the final checkpoint, whereas $k=32$ achieves the highest final accuracy of 84.31\%. Larger supports provide no further gains, suggesting that moderate support sizes offer the best balance between accuracy and late-stage retention.

\begin{figure*}[t]
  \centering
  \includegraphics[width=0.495\textwidth]{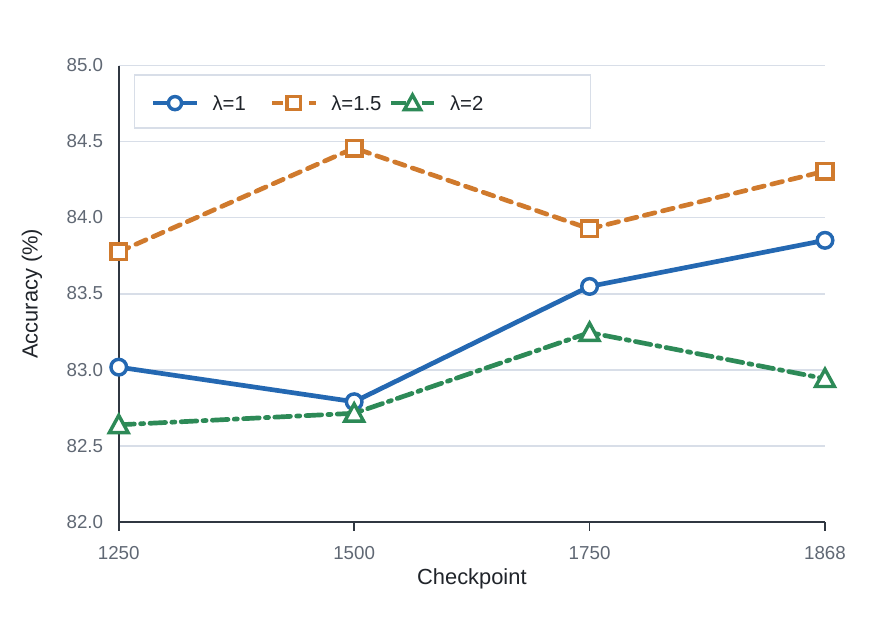}
  \hfill
  \includegraphics[width=0.495\textwidth]{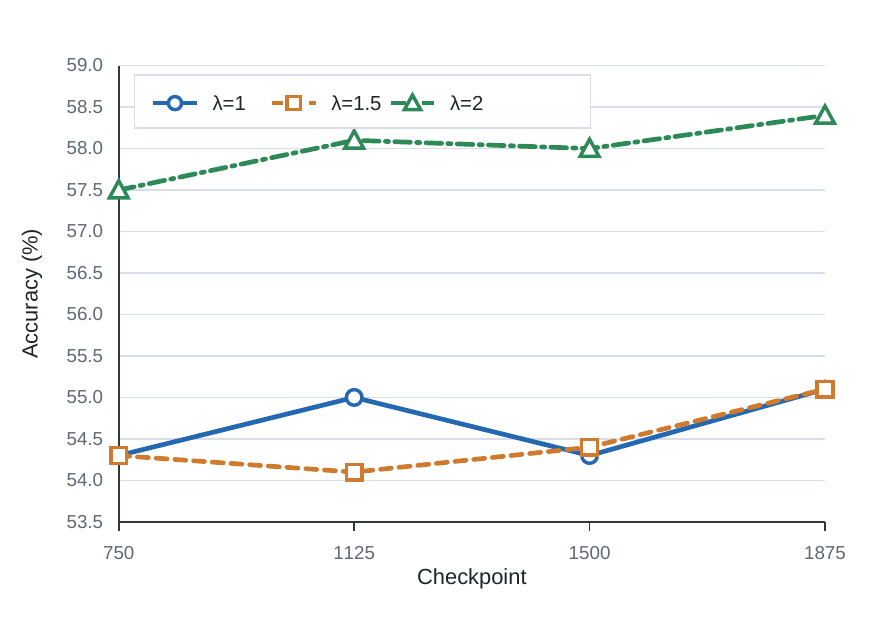}
  \caption{Training dynamics under different matching strengths $\lambda$. Left: GSM8K. Right: MATH. The preferred strength is task-dependent: $\lambda=1.5$ is strongest on GSM8K, while $\lambda=2$ is strongest on MATH.}
  \label{fig:matching-strength-dynamics}
\end{figure*}

Figure \ref{fig:matching-strength-dynamics} further shows that the preferred matching strength is task- and configuration-dependent. Across the evaluated checkpoints, $\lambda=1.5$ performs best on GSM8K and finishes at 84.31\%, while $\lambda=2$ leads on MATH and finishes at 58.4\%, compared with 55.1\% for $\lambda=1$ and $1.5$. A possible explanation is reward saturation: more frequent reward ties on GSM8K may make the advantage-based matching weights less selective, causing a larger $\lambda$ to amplify weakly differentiated rollout targets.

\section{Analysis and Discussion}

\subsection{Latent Thought Credit Diagnostics}
\label{sec:latent-thought-credit-diagnostics}

\subsubsection{Fixed-Context Probe Protocol}

To isolate thought-level task utility from answer-sampling noise, we use a fixed-context diagnostic that is not used for training. For each prompt, we sample $K$ latent thoughts, freeze each post-thought context $s_i$, and generate $M$ answers from that same context. We evaluate the pre-RL initial policy and the final LTC policy under the same continuous latent-thought rollout and Gumbel exploration settings.

Let the expected reward of the $i$-th latent thought be
\[
\mu_i=\mathbb{E}[r\mid x,s_i],
\]
where $r$ is the answer reward. We approximate $\mu_i$ with a held-out answer pool $\mu_i^{\mathrm{ref}}$ and compare it with low-budget estimates $\hat{\mu}_i(m)$ computed from $m$ probe answers. For the main diagnostic, we use 256 GSM8K test prompts, sample $K=4$ latent thoughts per prompt, and generate $M=40$ fixed-context answers per thought. Answers $0,\ldots,7$ form the probe pool, while answers $8,\ldots,39$ form the held-out reference pool, allowing $m\in\{1,2,4,8\}$.

\subsubsection{Variance Evidence and Budget Sensitivity}

We first compare between-thought expected-reward variance with within-thought answer variance. The former measures the spread in held-out mean correctness across latent thoughts for the same prompt, whereas the latter measures answer-sampling noise after the context $s_i$ is fixed. Table \ref{tab:fixed-context-variance} shows that within-thought variance exceeds between-thought variance in all three settings. For the initial policy, the noise-to-signal ratio is 1.54, showing that answer-sampling noise already exceeds thought-level variation before RL training. Compared with the initial policy, the final policy has lower between-thought variance (0.0043 vs.\ 0.0244) while retaining substantial within-thought variance (0.0568 vs.\ 0.0376), yielding a ratio of 13.10. This observed shift indicates that reliable thought-level credit estimation is more statistically demanding when candidate thoughts have closer expected utilities. Under a conditional-independence approximation, averaging over $m$ answers reduces the answer-sampling noise variance approximately to $\mathrm{within}/m$, consistent with the shared-context variance analysis of GRPO-MA \citep{wang2026treestylebranchingmattersthought}.

\begin{table}[t]
  \centering
  \small
  \renewcommand{\arraystretch}{1.08}
  \setlength{\tabcolsep}{2.5pt}
  \begin{tabular}{lrrrrrr}
    \toprule
    Setting & Prompts & $K$ & $M$ & Between & Within & Ratio \\
    \midrule
    Initial policy & 256 & 4 & 40 & 0.0244 & 0.0376 & 1.54 \\
    Final policy & 256 & 4 & 40 & 0.0043 & 0.0568 & 13.10 \\
    Final policy, high-$K$ & 256 & 32 & 64 & 0.0065 & 0.0583 & 8.99 \\
    \bottomrule
  \end{tabular}
  \caption{Fixed-context correctness-variance diagnostics. Between is the variance over held-out thought-level mean correctness; within is the answer-level correctness variance under fixed contexts; ratio is within / between.}
  \label{tab:fixed-context-variance}
\end{table}

With a broader candidate pool of $K=32$ latent thoughts and $M=64$ answers per thought, the ratio remains 8.99. Its decrease relative to the final-policy $K=4$ setting reflects the larger between-thought variance exposed by the broader candidate pool, while within-thought variance remains similar.

\subsubsection{Estimator Error}

Using the main diagnostic setting, we directly evaluate the discrepancy between the low-budget estimate $\hat{\mu}_i(m)$ and the held-out reference expected reward $\mu_i^{\mathrm{ref}}$. Table \ref{tab:fixed-context-estimator-error} shows that increasing the probe-answer budget consistently reduces estimation error for both policies. From $m=1$ to $m=8$, MAE decreases from 0.0785 to 0.0349 for the initial policy and from 0.1148 to 0.0490 for the final policy. The corresponding MSE values decrease from 0.0412 to 0.0068 and from 0.0581 to 0.0093, respectively. These results show that multi-answer averaging stabilizes thought-level expected-reward estimation both before and after training.

\begin{table}[t]
  \centering
  \small
  \renewcommand{\arraystretch}{1.08}
  \setlength{\tabcolsep}{3pt}
  \begin{tabular}{rcccc}
    \toprule
    & \multicolumn{2}{c}{Initial policy}
    & \multicolumn{2}{c}{Final policy} \\
    \cmidrule(lr){2-3}\cmidrule(lr){4-5}
    $m$ & MAE $\downarrow$ & MSE $\downarrow$
    & MAE $\downarrow$ & MSE $\downarrow$ \\
    \midrule
    1 & 0.0785 & 0.0412 & 0.1148 & 0.0581 \\
    2 & 0.0603 & 0.0219 & 0.0899 & 0.0324 \\
    4 & 0.0448 & 0.0112 & 0.0673 & 0.0180 \\
    8 & 0.0349 & 0.0068 & 0.0490 & 0.0093 \\
    \bottomrule
  \end{tabular}
  \caption{Estimator error in the main fixed-context diagnostic. MAE and MSE compare $\hat{\mu}_i(m)$ with the held-out reference mean for each latent thought.}
  \label{tab:fixed-context-estimator-error}
\end{table}

\subsubsection{Credit Assignment Consequence}

Finally, we examine whether answer-reward estimation error affects thought-level credit assignment. Table \ref{tab:tie-aware-credit} reports two fixed-context credit diagnostics relative to held-out reference utilities. Pairwise error assigns a penalty of 1 to reversed orderings and 0.5 to estimate-side ties, measuring incorrect or ambiguous thought ordering. Regret is the held-out reward gap between the selected thought and the best available thought, measuring the utility cost of noisy selection.

Increasing $m$ reduces both pairwise error and regret for the initial and final policies. For the initial policy, pairwise error decreases from 0.3840 to 0.2621 and regret decreases from 0.0406 to 0.0166. For the final policy, pairwise error remains higher but decreases from 0.4728 to 0.4380, while regret decreases from 0.0320 to 0.0254. These results show that multi-answer estimation improves thought-level credit assignment both before and after training.

\begin{table}[t]
  \centering
  \small
  \renewcommand{\arraystretch}{1.08}
  \setlength{\tabcolsep}{3pt}
  \begin{tabular}{lrrrr}
    \toprule
    Policy and metric & $m=1$ & $m=2$ & $m=4$ & $m=8$ \\
    \midrule
    Initial: pairwise error & 0.3840 & 0.3510 & 0.3150 & 0.2621 \\
    Initial: regret & 0.0406 & 0.0329 & 0.0233 & 0.0166 \\
    Final: pairwise error & 0.4728 & 0.4721 & 0.4450 & 0.4380 \\
    Final: regret & 0.0320 & 0.0302 & 0.0267 & 0.0254 \\
    \bottomrule
  \end{tabular}
  \caption{Fixed-context credit diagnostics under different probe-answer budgets. Lower pairwise error and regret indicate more reliable thought-level credit assignment.}
  \label{tab:tie-aware-credit}
\end{table}

\section{Conclusion and Future Work}

This paper introduced LTC, a hierarchical credit-assignment framework for latent reasoning. Instead of reinforcing a latent thought from a single sampled answer, LTC fixes the post-thought context, estimates thought-level expected reward by averaging multiple downstream answers, and combines multi-answer thought-level advantages with group-relative answer advantages and a thought-matching auxiliary loss. Experiments show that LTC achieves improved average performance among the compared methods, while fixed-context diagnostics show that multi-answer averaging reduces thought-level reward-estimation and credit-ordering errors. Together, these results support multi-answer credit assignment for latent reasoning. The current study focuses on Qwen2.5-3B/7B-Instruct and on mathematical reasoning and STEM multiple-choice benchmarks with verifiable or easily normalized rewards, so broader model families, multilingual and dialogue settings, noisier open-ended rewards, adaptive choices of $K$ and $M$, and interpretability of continuous latent thoughts remain important directions for future work.

\bibliography{custom}

@article{shao2024deepseekmath,
  title={DeepSeekMath: Pushing the Limits of Mathematical Reasoning in Open Language Models},
  author={Shao, Zhihong and Wang, Peiyi and Zhu, Qihao and Xu, Runxin and Song, Junxiao and Bi, Xiao and Zhang, Haowei and Zhang, Mingchuan and Li, Y. K. and Wu, Y. and Guo, Daya},
  journal={arXiv preprint arXiv:2402.03300},
  year={2024}
}

@article{hao2024training,
  title={Training large language models to reason in a continuous latent space},
  author={Hao, Shibo and Sukhbaatar, Sainbayar and Su, DiJia and Li, Xian and Hu, Zhiting and Weston, Jason and Tian, Yuandong},
  journal={arXiv preprint arXiv:2412.06769},
  year={2024}
}

@article{dou2025plan,
  title={Plan Then Action: High-Level Planning Guidance Reinforcement Learning for LLM Reasoning},
  author={Dou, Zhihao and Zhao, Qinjian and Wan, Zhongwei and Zhang, Dinggen and Wang, Weida and Raiyan, Towsif and Chen, Benteng and Pan, Qingtao and Ouyang, Yang and Song, Chaoda and Gao, Zhiqiang and Zhang, Shufei and Biswas, Sumon},
  journal={arXiv preprint arXiv:2510.01833},
  year={2025}
}

@misc{wang2026treestylebranchingmattersthought,
      title={Why Tree-Style Branching Matters for Thought Advantage Estimation in GRPO}, 
      author={Hongcheng Wang and Yinuo Huang and Sukai Wang and Guanghui Ren and Hao Dong},
      year={2026},
      eprint={2509.24494},
      archivePrefix={arXiv},
      primaryClass={cs.CL},
      url={https://arxiv.org/abs/2509.24494}, 
}

@inproceedings{yue2025hybrid,
  title={Hybrid Latent Reasoning via Reinforcement Learning},
  author={Yue, Zhenrui and Jin, Bowen and Zeng, Huimin and Zhuang, Honglei and Qin, Zhen and Yoon, Jinsung and Shang, Lanyu and Han, Jiawei and Wang, Dong},
  booktitle={Advances in Neural Information Processing Systems},
  volume={38},
  pages={5501--5530},
  publisher={Curran Associates, Inc.},
  year={2025},
  url={https://proceedings.neurips.cc/paper_files/paper/2025/file/087f7678af2abbe0c37ecc30bd7d1adf-Paper-Conference.pdf}
}

@article{zheng2025soft,
  title={Soft-grpo: Surpassing discrete-token llm reinforcement learning via gumbel-reparameterized soft-thinking policy optimization},
  author={Zheng, Zhi and Gu, Yu and Liu, Wei and Teh, Yee Whye and Lee, Wee Sun},
  journal={arXiv preprint arXiv:2511.06411},
  year={2025}
}

@article{williams2026prioritize,
  title={Prioritize the Process, Not Just the Outcome: Rewarding Latent Thought Trajectories Improves Reasoning in Looped Language Models},
  author={Williams, Jonathan and Tureci, Esin},
  journal={arXiv preprint arXiv:2602.10520},
  year={2026}
}

@misc{zhu2025surveylatentreasoning,
      title={A Survey on Latent Reasoning}, 
      author={Rui-Jie Zhu and Tianhao Peng and Tianhao Cheng and Xingwei Qu and Jinfa Huang and Dawei Zhu and Hao Wang and Kaiwen Xue and Xuanliang Zhang and Yong Shan and Tianle Cai and Taylor Kergan and Assel Kembay and Andrew Smith and Chenghua Lin and Binh Nguyen and Yuqi Pan and Yuhong Chou and Zefan Cai and Zhenhe Wu and Yongchi Zhao and Tianyu Liu and Jian Yang and Wangchunshu Zhou and Chujie Zheng and Chongxuan Li and Yuyin Zhou and Zhoujun Li and Zhaoxiang Zhang and Jiaheng Liu and Ge Zhang and Wenhao Huang and Jason Eshraghian},
      year={2025},
      eprint={2507.06203},
      archivePrefix={arXiv},
      primaryClass={cs.CL},
      url={https://arxiv.org/abs/2507.06203}, 
}

@article{tang2026multiplex,
  title={Multiplex Thinking: Reasoning via Token-wise Branch-and-Merge},
  author={Tang, Yao and Dong, Li and Hao, Yaru and Dong, Qingxiu and Wei, Furu and Gu, Jiatao},
  journal={arXiv preprint arXiv:2601.08808},
  year={2026}
}

@article{su2025token,
  title={Token assorted: Mixing latent and text tokens for improved language model reasoning},
  author={Su, DiJia and Zhu, Hanlin and Xu, Yingchen and Jiao, Jiantao and Tian, Yuandong and Zheng, Qinqing},
  journal={arXiv preprint arXiv:2502.03275},
  year={2025}
}

@misc{geiping2025scalingtesttimecomputelatent,
      title={Scaling up Test-Time Compute with Latent Reasoning: A Recurrent Depth Approach}, 
      author={Jonas Geiping and Sean McLeish and Neel Jain and John Kirchenbauer and Siddharth Singh and Brian R. Bartoldson and Bhavya Kailkhura and Abhinav Bhatele and Tom Goldstein},
      year={2025},
      eprint={2502.05171},
      archivePrefix={arXiv},
      primaryClass={cs.LG},
      url={https://arxiv.org/abs/2502.05171}, 
}

@misc{li2025seekdarkreasoningtesttime,
      title={Seek in the Dark: Reasoning via Test-Time Instance-Level Policy Gradient in Latent Space}, 
      author={Hengli Li and Chenxi Li and Tong Wu and Xuekai Zhu and Yuxuan Wang and Zhaoxin Yu and Eric Hanchen Jiang and Song-Chun Zhu and Zixia Jia and Ying Nian Wu and Zilong Zheng},
      year={2025},
      eprint={2505.13308},
      archivePrefix={arXiv},
      primaryClass={cs.LG},
      url={https://arxiv.org/abs/2505.13308}, 
}

@misc{wei2023chainofthoughtpromptingelicitsreasoning,
      title={Chain-of-Thought Prompting Elicits Reasoning in Large Language Models}, 
      author={Jason Wei and Xuezhi Wang and Dale Schuurmans and Maarten Bosma and Brian Ichter and Fei Xia and Ed Chi and Quoc Le and Denny Zhou},
      year={2023},
      eprint={2201.11903},
      archivePrefix={arXiv},
      primaryClass={cs.CL},
      url={https://arxiv.org/abs/2201.11903}, 
}

@misc{wang2023selfconsistencyimproveschainthought,
      title={Self-Consistency Improves Chain of Thought Reasoning in Language Models}, 
      author={Xuezhi Wang and Jason Wei and Dale Schuurmans and Quoc Le and Ed Chi and Sharan Narang and Aakanksha Chowdhery and Denny Zhou},
      year={2023},
      eprint={2203.11171},
      archivePrefix={arXiv},
      primaryClass={cs.CL},
      url={https://arxiv.org/abs/2203.11171}, 
}

@misc{yao2023treethoughtsdeliberateproblem,
      title={Tree of Thoughts: Deliberate Problem Solving with Large Language Models}, 
      author={Shunyu Yao and Dian Yu and Jeffrey Zhao and Izhak Shafran and Thomas L. Griffiths and Yuan Cao and Karthik Narasimhan},
      year={2023},
      eprint={2305.10601},
      archivePrefix={arXiv},
      primaryClass={cs.CL},
      url={https://arxiv.org/abs/2305.10601}, 
}

@misc{cobbe2021trainingverifierssolvemath,
      title={Training Verifiers to Solve Math Word Problems}, 
      author={Karl Cobbe and Vineet Kosaraju and Mohammad Bavarian and Mark Chen and Heewoo Jun and Lukasz Kaiser and Matthias Plappert and Jerry Tworek and Jacob Hilton and Reiichiro Nakano and Christopher Hesse and John Schulman},
      year={2021},
      eprint={2110.14168},
      archivePrefix={arXiv},
      primaryClass={cs.LG},
      url={https://arxiv.org/abs/2110.14168}, 
}

@misc{lightman2023letsverifystepstep,
      title={Let's Verify Step by Step}, 
      author={Hunter Lightman and Vineet Kosaraju and Yura Burda and Harri Edwards and Bowen Baker and Teddy Lee and Jan Leike and John Schulman and Ilya Sutskever and Karl Cobbe},
      year={2023},
      eprint={2305.20050},
      archivePrefix={arXiv},
      primaryClass={cs.LG},
      url={https://arxiv.org/abs/2305.20050}, 
}

@misc{jang2017categoricalreparameterizationgumbelsoftmax,
      title={Categorical Reparameterization with Gumbel-Softmax}, 
      author={Eric Jang and Shixiang Gu and Ben Poole},
      year={2017},
      eprint={1611.01144},
      archivePrefix={arXiv},
      primaryClass={stat.ML},
      url={https://arxiv.org/abs/1611.01144}, 
}

@misc{maddison2017concretedistributioncontinuousrelaxation,
      title={The Concrete Distribution: A Continuous Relaxation of Discrete Random Variables}, 
      author={Chris J. Maddison and Andriy Mnih and Yee Whye Teh},
      year={2017},
      eprint={1611.00712},
      archivePrefix={arXiv},
      primaryClass={cs.LG},
      url={https://arxiv.org/abs/1611.00712}, 
}

@misc{hendrycks2021measuringmathematicalproblemsolving,
      title={Measuring Mathematical Problem Solving With the MATH Dataset}, 
      author={Dan Hendrycks and Collin Burns and Saurav Kadavath and Akul Arora and Steven Basart and Eric Tang and Dawn Song and Jacob Steinhardt},
      year={2021},
      eprint={2103.03874},
      archivePrefix={arXiv},
      primaryClass={cs.LG},
      url={https://arxiv.org/abs/2103.03874}, 
}

@misc{hendrycks2021measuringmassivemultitasklanguage,
      title={Measuring Massive Multitask Language Understanding}, 
      author={Dan Hendrycks and Collin Burns and Steven Basart and Andy Zou and Mantas Mazeika and Dawn Song and Jacob Steinhardt},
      year={2021},
      eprint={2009.03300},
      archivePrefix={arXiv},
      primaryClass={cs.CY},
      url={https://arxiv.org/abs/2009.03300}, 
}

@misc{clark2018thinksolvedquestionanswering,
      title={Think you have Solved Question Answering? Try ARC, the AI2 Reasoning Challenge}, 
      author={Peter Clark and Isaac Cowhey and Oren Etzioni and Tushar Khot and Ashish Sabharwal and Carissa Schoenick and Oyvind Tafjord},
      year={2018},
      eprint={1803.05457},
      archivePrefix={arXiv},
      primaryClass={cs.AI},
      url={https://arxiv.org/abs/1803.05457}, 
}

@article{qwen2024qwen25technicalreport,
      title={Qwen2.5 Technical Report}, 
      author={Yang, An and Yang, Baosong and Zhang, Beichen and Hui, Binyuan and Zheng, Bo and Yu, Bowen and Li, Chengyuan and Liu, Dayiheng and Huang, Fei and Wei, Haoran and Lin, Huan and Yang, Jian and Tu, Jianhong and Zhang, Jianwei and Yang, Jianxin and Yang, Jiaxi and Zhou, Jingren and Lin, Junyang and Dang, Kai and Lu, Keming and Bao, Keqin and Yang, Kexin and Yu, Le and Li, Mei and Xue, Mingfeng and Zhang, Pei and Zhu, Qin and Men, Rui and Lin, Runji and Li, Tianhao and Xia, Tingyu and Ren, Xingzhang and Ren, Xuancheng and Fan, Yang and Su, Yang and Zhang, Yichang and Wan, Yu and Liu, Yuqiong and Cui, Zeyu and Zhang, Zhenru and Qiu, Zihan},
      journal={CoRR},
      volume={abs/2412.15115},
      year={2024},
      doi={10.48550/arXiv.2412.15115},
      url={https://doi.org/10.48550/arXiv.2412.15115}
}

@misc{chen2021evaluatinglargelanguagemodels,
      title={Evaluating Large Language Models Trained on Code}, 
      author={Mark Chen and Jerry Tworek and Heewoo Jun and Qiming Yuan and Henrique Ponde de Oliveira Pinto and Jared Kaplan and Harri Edwards and Yuri Burda and Nicholas Joseph and Greg Brockman and Alex Ray and Raul Puri and Gretchen Krueger and Michael Petrov and Heidy Khlaaf and Girish Sastry and Pamela Mishkin and Brooke Chan and Scott Gray and Nick Ryder and Mikhail Pavlov and Alethea Power and Lukasz Kaiser and Mohammad Bavarian and Clemens Winter and Philippe Tillet and Felipe Petroski Such and Dave Cummings and Matthias Plappert and Fotios Chantzis and Elizabeth Barnes and Ariel Herbert-Voss and William Hebgen Guss and Alex Nichol and Alex Paino and Nikolas Tezak and Jie Tang and Igor Babuschkin and Suchir Balaji and Shantanu Jain and William Saunders and Christopher Hesse and Andrew N. Carr and Jan Leike and Josh Achiam and Vedant Misra and Evan Morikawa and Alec Radford and Matthew Knight and Miles Brundage and Mira Murati and Katie Mayer and Peter Welinder and Bob McGrew and Dario Amodei and Sam McCandlish and Ilya Sutskever and Wojciech Zaremba},
      year={2021},
      eprint={2107.03374},
      archivePrefix={arXiv},
      primaryClass={cs.LG},
      url={https://arxiv.org/abs/2107.03374}, 
}

\end{document}